\documentclass[10pt,twocolumn,letterpaper]{article}

\usepackage{cvpr}              

\usepackage{multirow}
\usepackage{makecell}
\usepackage{array}
\usepackage{pifont}
\usepackage{float}
\newcolumntype{Y}{>{\centering\arraybackslash}p{1.25cm}}

\usepackage[table]{xcolor}
\newcommand{\gcell}{\cellcolor{gray!15}}

\definecolor{cvprblue}{rgb}{0.21,0.49,0.74}
\usepackage[pagebackref,breaklinks,colorlinks,allcolors=cvprblue]{hyperref}

\def\paperID{}
\def\confName{CVPR}
\def\confYear{2026}

\title{ImmerIris: A Large-Scale Dataset and Benchmark for Off-Axis and Unconstrained Iris Recognition in Immersive Applications}

\author{
    Yuxi Mi$^{1}$\thanks{Authors contributed equally to this paper.}\quad
    Qiuyang Yuan$^{1}\footnotemark[1]$\quad
    Zhizhou Zhong$^{1}$\quad
    Xuan Zhao$^{1}$\\
    Jiaogen Zhou$^{2}$\thanks{Corresponding authors.}\quad
    Fubao Zhu$^{3}\footnotemark[2]$\quad
    Jihong Guan$^{4}$\quad
    Shuigeng Zhou$^{1}\footnotemark[2]$
    \\
    $^{1}$Shanghai Key Lab of Intelligent Information Processing, Fudan University \\
    $^{2}$Huaiyin Normal University \quad
    $^{3}$Zhengzhou University of Light Industry \quad
    $^{4}$Tongji University 
    \\
    {\tt\small \{yxmi20, sgzhou\}@fudan.edu.cn, \{qyyuan23, zzzhong22, xzhao23\}@m.fudan.edu.cn} \\
    {\tt\small zhoujg@hytc.edu.cn, fbzhu@zzuli.edu.cn, jhguan@tongji.edu.cn}
}

\begin{document}
\maketitle

\begin{abstract}

Recently, iris recognition is regaining prominence in immersive applications such as extended reality as a means of seamless user identification. This application scenario introduces unique challenges compared to traditional iris recognition under controlled setups, as the ocular images are primarily captured off-axis and less constrained, causing perspective distortion, intra-subject variation, and quality degradation in iris textures. Datasets capturing these challenges remain limited. This paper fills this gap by presenting a large-scale iris dataset collected via head-mounted displays, termed \textit{ImmerIris}. It contains 499,791 ocular images from 546 subjects, and is, to our knowledge, the largest public iris dataset to date and among the first dedicated to immersive applications. It is accompanied by a comprehensive set of evaluation protocols that benchmark recognition systems under various challenging conditions. This paper also draws attention to a shared obstacle of current recognition methods, the reliance on a pre-processing, normalization stage, which is fallible in off-axis and unconstrained setups. To this end, this paper further proposes a normalization-free paradigm that directly learns from minimally adjusted ocular images. Despite its simplicity, it outperforms normalization-based prior arts, indicating a promising direction for robust iris recognition.


\vspace{-3mm}
\end{abstract}
\section{Introduction}
\label{sec:intro}

\begin{figure}
    \centering
    \includegraphics[width=\linewidth]{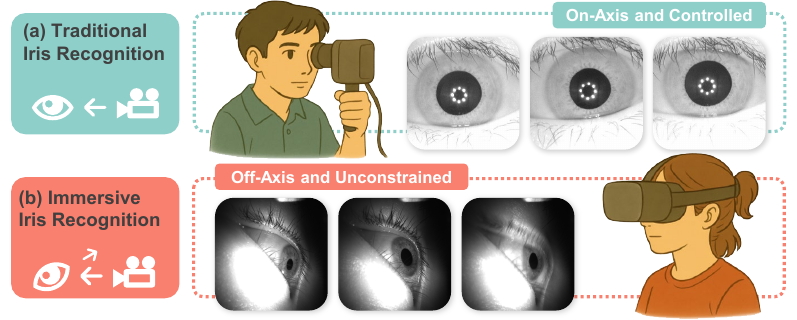}
    \caption{Comparison of application scenarios. Each sample group is from the same person. (a) Traditional iris recognition acquires on-axis and controlled images with dedicated devices, with samples being highly invariant. (b) Immersive iris recognition collects images using consumer HMDs, yielding off-axis and unconstrained samples that exhibit distortion, variation, and degradation.}
    \label{fig:immer-demo}
    \vspace{-5mm}
\end{figure}

Iris recognition is a long-standing biometric technique to identify persons by the distinctive textures of iris, the circular, light-regulating structure in human eyes. Traditionally, iris recognition is used for sensitive institutional applications such as border control, where users gaze at specialized cameras to acquire on-axis ocular images in a controlled setup~\cite{nguyen2024deep} (\cref{fig:immer-demo}(a)). With the recent rise of immersive applications such as extended reality (XR), \textit{immersive iris recognition} has gained prominence~\cite{wang2022human}, as ocular images can now be easily acquired through consumer head-mounted displays (HMD) like virtual reality (VR) headsets to seamlessly recognize users for tasks including login and e-payment. To comply with hardware design and user experience, immersive iris recognition usually differs significantly in image acquisition: HMD cameras are usually placed at an angle to eyes (\ie, with users not gazing at the camera), producing \textit{off-axis} images. The images are also captured in \textit{unconstrained} environments where conditions vary and users may not be fully cooperative (\cref{fig:immer-demo}(b)).

Data scarcity has long been a barrier to iris recognition research~\cite{omelina2021survey,chun2004iris,casia-iris-v4}, where most existing datasets are either proprietary or small in scale. In immersive iris recognition, this scarcity is further exacerbated, as the off-axis and unconstrained acquisition introduces three unique challenges compared to the controlled setup, namely: \textit{Perspective distortion}, where tilted camera-eye geometries make the iris appear elliptical and unevenly stretched in local textures; \textit{Intra-subject variation}, where irises from the same eye may vary noticeably, arising from environmental and user behavioral changes, \eg, variations in illumination and gaze direction; \textit{Quality degradation}, where non-cooperative user interaction can yield low-quality samples, \eg, occluded irises when the eyes are partially closed. To date, datasets capturing all three challenges remain very limited.

This paper fills the gap by presenting \textit{ImmerIris}, a large-scale public iris dataset acquired off-axis and unconstrained via VR headsets. It consists of 499,791 ocular images from 546 subjects, and is, to our knowledge, the largest public iris dataset and among the first dedicated to immersive applications. This paper also organizes ImmerIris into a comprehensive set of evaluation protocols to benchmark recognition systems under varying operation modes and challenging conditions. The dataset and benchmark are expected to advance research viability on immersive iris recognition.

\begin{figure}
    \centering
    \includegraphics[width=\linewidth]{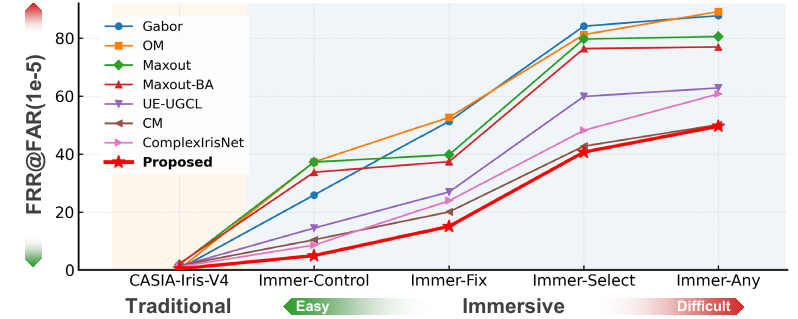}
    \caption{Performance of SOTAs and our normalization-free approach on CASIA-Iris-V4~\cite{casia-iris-v4} and 4 increasingly challenging ImmerIris protocols. Lower FRR is better. SOTAs perform well on the traditional setup yet drop sharply under immersive conditions, whereas our approach consistently outperforms them.}
    \label{fig:intro-comp}
    \vspace{-3mm}
\end{figure}

Based on the dataset, this paper investigates state-of-the-art (SOTA) recognition methods~\cite{daugman2009iris,zhang2018deep,wei2022towards,wei2022contextual,nguyen2022complex} designed for the controlled setup and finds that they all generalize \textit{unsatisfactorily} to the immersive scenario. In experiments, SOTAs are trained and tested both on a controlled iris dataset~\cite{casia-iris-v4} and on ImmerIris under four evaluation protocols of increasing difficulty, as detailed in~\cref{sec:dataset,sec:benchmark}. In~\cref{fig:intro-comp}, a very low false rejection rate (FRR) is first observed on the controlled dataset, where SOTAs demonstrate their proven success. However, FRR increases sharply on ImmerIris protocols, revealing a substantial performance drop. Architecturally, most SOTAs share a common two-stage paradigm: a \textit{normalization} stage that unwraps the iris region into a rectangular strip of normalized texture, followed by a \textit{feature extraction} stage that maps the texture into identity templates. This paper finds that the normalization stage imposes a major bottleneck, as current techniques often yield ill-unwrapped textures under immersive distortion and degradation, and fail to handle intra-subject variations. Though some SOTAs~\cite{wei2022towards,wei2022contextual,nguyen2022complex} post-process the normalized textures with more adaptive designs, such approaches are largely non-intuitive and suboptimal.

To improve recognizability in the immersive scenario, this paper reframes the paradigm into an end-to-end one that dispenses with the fallible normalization and directly learns from minimally adjusted ocular images. Simply, it crops the iris region from ocular images using a reliably obtained bounding box. The cropped region preserves both iris texture and surrounding contextual cues. For feature extraction, it adopts the proven practice of modern face recognition (FR) systems, whose success lies not in elaborate preprocessing but rather in robust feature extractors and discriminative objectives. In~\cref{fig:intro-comp} and~\cref{sec:benchmark}, this simple yet intuitive design is found to outperform normalization-based SOTAs in most cases, highlighting a promising direction for robust immersive iris recognition.

This paper makes the following three main contributions:
\begin{enumerate}
\item We introduce \textit{ImmerIris}, a large-scale, off-axis and unconstrained iris dataset for immersive applications.
\item We establish a comprehensive set of evaluation protocols based on ImmerIris, and demonstrate that SOTAs are not readily generalizable to the immersive scenario.
\item We propose a simple yet effective iris recognition paradigm that dispenses with the fallible normalization and achieves superior performance in most cases.
\end{enumerate}

\section{Related Work}
\label{sec:rw}

\begin{table}[tbp]

\caption{Comparison of existing iris recognition datasets and ImmerIris in terms of acquisition setup and scale.}
\label{tab:rw-dataset}

\centering
\footnotesize
\begin{tabular}{lccc}
\toprule
\textbf{Dataset} & \textbf{Axis} & \textbf{Conditions} & \textbf{\# Img (Subject)} \\
\midrule
CASIA-IrisV1~\cite{casia-iris-v1} & On    & Controlled        & 36,240 (50)                 \\
CASIA-IrisV4~\cite{casia-iris-v4}  & On  & Controlled        & 20,000 (1000)                \\
IITD-V1~\cite{kumar2010comparison}  & On       & Controlled        & 1,120 (112)                  \\
CUHK Iris~\cite{chun2004iris} & On      & Controlled        & 254 (18)                     \\
ND-CrossSensor~\cite{xiao2013coupled} & On  & Controlled        & 117,503 (676)               \\
UBIRIS.V1~\cite{proencca2005ubiris} & On   & Semi-Ctrl         & 1,877 (241)               \\
UBIRIS.V2~\cite{proencca2009ubiris} & On      & Semi-Ctrl         & 11,102 (261)                 \\
VISOB~\cite{rattani2016icip}  & On         & Semi-Ctrl         & 95,107 (550)               \\
CASIA-BTAS~\cite{zhang2016btas}  & On    & Semi-Ctrl         & 4,500 (150)                  \\
PolyU Iris DB~\cite{wang2022human}  & Off    & Semi-Ctrl         & 142,005 (384)                  \\
\rowcolor{gray!15}
\textbf{ImmerIris} (ours)  & Off     & Unconstrained        & 499,791 (546)          \\
\bottomrule
\end{tabular}
\vspace{-3mm}
\end{table}

\subsection{Iris Recognition Datasets}
\label{subsec:rw-ird}

Iris recognition datasets consist of ocular images acquired using visible-light (VIS) or near-infrared (NIR) cameras~\cite{omelina2021survey}. Pioneering datasets such as CASIA-IrisV1~\cite{casia-iris-v1}, CASIA-IrisV4~\cite{casia-iris-v4}, the IIT Delhi database~\cite{kumar2010comparison}, the CUHK Iris dataset~\cite{chun2004iris}, and ND-CrossSensor~\cite{arora2012iris,xiao2013coupled} are collected on-axis (\ie, with cameras frontal to gaze axis) in controlled laboratory setups. Later studies introduce datasets with certain acquisition conditions intentionally set non-ideal, such as using non-professional devices~\cite{rattani2016icip,zhang2016btas}, capturing at a distance~\cite{proencca2009ubiris}, or including occlusions~\cite{proencca2005ubiris}. Nonetheless, they still assume semi-controlled environments and cooperative users (\eg, fixed gaze and illumination), with all images captured on-axis, hence fall short of covering the unique challenges posed by immersive scenarios.
Most recently, the PolyU Iris DB~\cite{wang2022human} employs a VR device for acquisition, which is conceptually close to ImmerIris. However, ImmerIris further investigates real-world challenges such as pronounced off-axis distortions and illumination changes. \Cref{tab:rw-dataset} summarizes a shortlist of datasets. A full summary is deferred to the supplementary material.

\subsection{Iris Recognition Methods}
\label{subsec:rw-ir}

Current iris recognition methods mainly follow a two-stage paradigm~\cite{nguyen2024deep}. It begins with a \textit{normalization} stage that segments the iris region from ocular images~\cite{daugman2002high,he2008toward,vatsa2008improving}, parameterizes the iris contour~\cite{shah2009iris,proenca2009iris}, and unwraps the contour into a rectangular strip of normalized textures via a polar transform~\cite{daugman2009iris}. A subsequent \textit{feature extraction} stage maps the normalized texture to distinctive identity templates. Here, early approaches employ hand-crafted filters such as Gabor~\cite{daugman2009iris}, log-Gabor~\cite{ali2007recognition}, ordinal measure~\cite{sun2008ordinal}, sparse representation~\cite{pillai2011secure}, and phase correlation~\cite{miyazawa2008effective} to produce binarized \textit{iriscodes}, which are majorly comparable via Hamming distance~\cite{norouzi2012hamming}. Recent methods leverage deep neural networks (DNN) with varied architectures, including CNNs~\cite{gangwar2016deepirisnet,nguyen2017iris,zhang2018deep,wei2022towards}, FCNs~\cite{zhao2017towards}, Mask R-CNN~\cite{zhao2019deep}, DenseNet~\cite{wang2019toward,boutros2020benchmarking}, ResNet~\cite{boutros2020benchmarking}, and backbones ensembling periocular cues~\cite{wei2022contextual,nguyen2022complex,zhao2018improving}. Some efforts~\cite{ahmad2019thirdeye,birgale2010iris,gangwar2019deepirisnet2} investigate normalization-free recognition and are conceptually close to our later method. Yet, they still rely on iris segmentation, a fallible part of the normalization pipeline. Overall, though current methods achieve remarkable success in controlled setups~\cite{nguyen2017long} and partly under semi-controlled imaging~\cite{wang2020recognition}, they are not designed for off-axis and unconstrained immersive iris recognition and perform unsatisfactorily, as later analyzed in~\cref{sec:baseline,sec:benchmark}.

\section{The ImmerIris Dataset}
\label{sec:dataset}

\subsection{Data Acquisition}
\label{subsec:data-acquisition}

\begin{figure}
    \centering
    \includegraphics[width=0.95\linewidth]{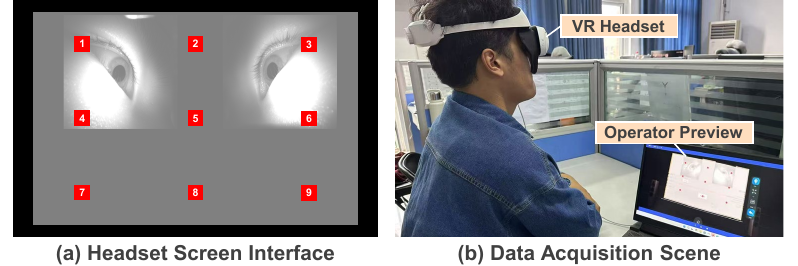}
    \caption{Data acquisition setup. (a) Screen interface of the VR headset, where red squares numbered 1-9 mark gaze points for sequential fixation. Live camera previews assist proper wearing. A full-screen white panel gradually increases in brightness to simulate illumination changes. (b) Actual scene of data acquisition.}
    \label{fig:acquisition}
    \vspace{-5mm}
\end{figure}

We collect NIR ocular images from human subjects using a general-purpose VR headset (Skyworth Pancake XR) equipped with custom-developed acquisition software. The headset features dual-eye displays to show instructions to the wearer and side-mounted cameras to capture off-axis images. The setup closely mimics real-world immersive XR experiences. All subjects are volunteers who provided informed consent, received no monetary compensation, and had no personal details recorded. Demographically, they are Asian adults aged 20-40 years, with a nearly balanced biological sex distribution. The data acquisition procedure was approved by the institutional review board (IRB).

Immersive iris recognition differs from the controlled setup in three distinctive challenges, as discussed in~\cref{sec:intro}. The first is \textit{perspective distortion}, which is deliberately reproduced in our acquisition setup using side-mounted cameras. Second, images from the same eye exhibit \textit{intra-subject variations} due to environmental and user behavioral changes, most notably in illumination and gaze direction: \textit{Illumination} often varies with screen brightness or ambient lighting, while \textit{gaze direction} changes depending on displayed visual content and user interaction. Our acquisition process captures these two factors by collecting ocular images with explicit variations from each subject.

Specifically, during acquisition, subjects wear the headset and view a 3$\times$3 grid of red squares numbered 1-9 on the screen, along with a live camera preview that assists proper device wearing, as shown in \cref{fig:acquisition}. The subjects are instructed to sequentially gaze and pause at each square to imitate real-world gaze variations. At each gaze point, the headset automatically adjusts the screen brightness across 11 levels, from the darkest to the brightest, to simulate illumination changes that affects imaging and induces natural pupil size variations. It captures 5 ocular images per eye with a resolution of 640$\times$640 at each level. In total, 990 images are captured per subject for both eyes combined under varied conditions. The dataset enrolls 546 subjects and initially comprises 540,540 ocular images. It will be publicly released to support future iris recognition researches.

\subsection{Data Cleaning and Annotation}
\label{subsec:data-cleaning}

\begin{figure}
    \centering
    \includegraphics[width=\linewidth]{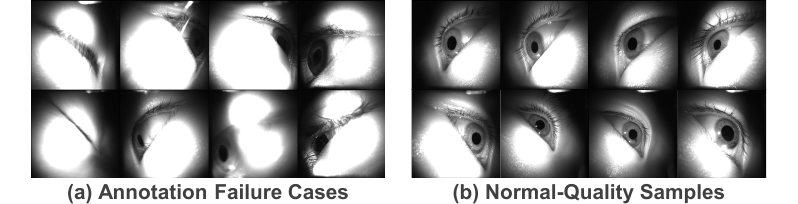}
    \caption{Sample ocular images. (a) Images that fail annotation are obviously flawed and therefore removed. (b) Normal samples.}
    \label{fig:cleaning}
    \vspace{-4mm}
\end{figure}

The third challenge of immersive iris recognition is image \textit{quality degradation}, arising from less well-calibrated devices (\eg, with suboptimal inter-pupillary distance or focal length) or non-cooperative user interactions (\eg, with partially closed eyes). In principle, we first remove obviously unrecognizable flawed samples, and annotate the remaining ones according to their potential types of degradation.

To remove severely flawed samples, we first employ a pretrained ocular detection model to obtain bounding boxes of the iris regions and discard 36,697 images that fail detection. \Cref{fig:cleaning}(a) shows exemplar failure cases, including ocular regions extending beyond the frame, closed eyes, and severe motion blur caused by blinking or gaze shifts. We further remove 4,052 images that are defective for iris recognition through visual inspection. After cleaning, 499,791 images remain. \Cref{fig:cleaning}(b) shows normal samples for comparison. 

\begin{figure}
    \centering
    \includegraphics[width=\linewidth]{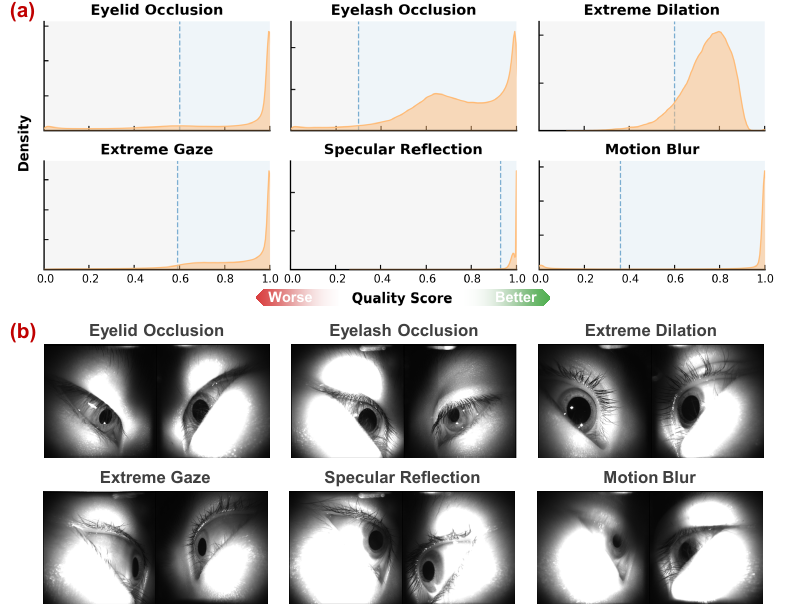}
    \caption{Data annotation. (a) Quality scores across 6 dimensions. Samples with quality scores below the thresholds (vertical lines) are considered degraded, and the rest as normal. (b) Examples from each quality-degradation category.}
    \label{fig:quality-distribution}
    \vspace{-3mm}
\end{figure}

Next, we annotate the remaining images with quality scores for 6 types of degradations that are frequently reported~\cite{omelina2021survey}: \textit{occlusion by eyelid}, \textit{occlusion by eyelash}, \textit{extensively dilated pupil}, \textit{extreme off-axis gaze}, \textit{specular reflection}, and \textit{motion blur}. Each quality dimension is thresholded to distinguish normal from degraded samples. \Cref{fig:quality-distribution}(a) presents the score distributions and corresponding thresholds. We find that approximately 42\% of images are degraded in at least one quality dimension, which further confirms that quality degradation poses a significant challenge in immersive iris recognition. \Cref{fig:quality-distribution}(b) shows samples for each type of degradation. See the supplementary material for implementation details and additional samples.

\subsection{Training-Test Partition}
\label{subsec:data-division}

We then partition the annotated images into a training set and a general test set with a ratio of 7:3. The training set comprises 347,927 images from 380 subjects, and the test set contains 151,864 images from 166 subjects. Subjects are non-overlapping between the two sets to ensure an open-set evaluation, where the recognition system is supposed to enroll and identify previously unseen persons. Following standard iris recognition practice~\cite{daugman2009iris}, we label two eyes from the same subject as two distinct classes. The test set is then divided into a comprehensive set of evaluation protocols.

\subsection{Protocol Design}
\label{subsec:protocol-rationale}

\begin{table}[tbp]
\caption{Summary of evaluation protocols by the factors studied. ``I, G, O, D, R, B'' denote illumination, gaze, occlusion, dilation, reflection, and blur. Symbols ``$\bullet$, $\circ$, $\triangle$'' indicate explicitly, partially, and implicitly included factors, and ``$\times$'' indicates exclusion.}
\label{tab:dataset-prop}

\centering
\footnotesize
\begin{tabular}{l|c|cc|cccc}
\toprule
\multirow{2}{*}{\textbf{Protocol}} &
\multirow{2}{*}{\textbf{Distortion}} &
\multicolumn{2}{c|}{\textbf{Variation}} &
\multicolumn{4}{c}{\textbf{Degradation}} \\
& & \textbf{I} & \textbf{G} &
  \textbf{O} & \textbf{D} &
  \textbf{R} & \textbf{B} \\
\midrule
Occlusion & $\bullet$ & $\triangle$ & $\times$ & $\bullet$ & $\times$ & $\times$ & $\times$ \\
Dilation  & $\bullet$ & $\triangle$ & $\times$ & $\times$ & $\bullet$ & $\times$ & $\times$ \\
Light     & $\bullet$ & $\bullet$ & $\times$ & $\times$ & $\triangle$ & $\times$ & $\times$ \\
Gaze      & $\bullet$ & $\triangle$ & $\bullet$ & $\times$ & $\times$ & $\times$ & $\times$ \\
Control   & $\bullet$ & $\triangle$ & $\times$ & $\times$ & $\times$ & $\times$ & $\times$ \\
Fix       & $\bullet$ & $\bullet$ & $\times$ & $\bullet$ & $\bullet$ & $\bullet$ & $\bullet$ \\
Select  & $\bullet$ & $\bullet$ & $\circ$ & $\bullet$ & $\bullet$ & $\bullet$ & $\bullet$ \\
Any       & $\bullet$ & $\bullet$ & $\bullet$ & $\bullet$ & $\bullet$ & $\bullet$ & $\bullet$ \\
\bottomrule
\end{tabular}
\vspace{-3mm}
\end{table}

We organize the types of intra-subject variations and quality degradations discussed above into 8 evaluation protocols that systematically examine their isolated and combined effects on iris recognition performance. Note that off-axis perspective distortion is inherent throughout the dataset and hence is not studied independently.

\noindent \textbf{Isolated effect.} We construct 4 protocols to study the isolated challenges posed by occlusion, dilated pupils, illumination change and gaze variation, respectively. Other factors (\ie, extreme gaze, reflection and motion blur) are later incorporated in the combined studies, as they occur sporadically throughout the dataset and are not easily isolable.

\begin{itemize}
    \item \textbf{Immer-Occlusion.} Partially closed eyes can cause eyelid and eyelash occlusion that obscure iris textures. These two cases are merged into a single protocol, as eyelash-occluded samples are found relatively scarce (3,664 in total). We select pairs with at least one occluded image, later during pair formation (\cref{subsec:protocol-organization}). To isolate its effect, gaze variation is minimized by treating images captured at the same gaze point (\ie, the same red square) as having the same gaze direction and pairing them accordingly, which also applies to the subsequent two protocols.
    \item \textbf{Immer-Dilation.} Extensive pupil dilation may result from individual physiological traits or dim illumination. It can compress surrounding iris textures and reduce recognizability. We select pairs where both images have low iris-to-ocular ratios based on the quality score.
    \item \textbf{Immer-Light.} Illumination changes affect both imaging and pupil size. We select images from 3 low brightness levels (preferably with dilation) and 3 high ones. It differs mainly from Immer-Dilation in that this protocol pairs dilated images with normal ones.
    \item \textbf{Immer-Gaze.} Gaze direction varies with the user's eye focus, altering the eye-camera geometry and hindering invariant feature extraction. It is later found to impact recognition performance most significantly. Here, we pair images captured at different gaze points and eliminate degradations by selecting only normal images.
\end{itemize}

\noindent \textbf{Combined effects.} We also organize 4 protocols to study the effects of multiple challenges in combination. They can be viewed as general evaluations of recognition systems under different operating modes: While immersive iris recognition is generally unconstrained, an operator may impose specific prerequisites (\eg, requiring users to fix gaze or remain cooperative) to balance convenience and reliability. Protocols are arranged in increasing difficulty:

\begin{itemize}
    \item \textbf{Immer-Control.} Users fix their gaze and be fully cooperative. This protocol differs from the traditional controlled setup only by the presence of off-axis distortion. We select normal images and pairing at the same gaze point.
    \item \textbf{Immer-Fix.} Users fix their gaze, but other conditions such as occlusion may vary. It involves both normal and degraded images, including reflection or blur.
    \item \textbf{Immer-Select.} Users are encouraged to avoid certain regions (\eg, the corners of view) that cause extreme gazes, while still being allowed to shift their gaze naturally. We empirically find that extreme gazes mostly occur at the innermost directions of each eye. Accordingly, we exclude images from these gaze points (right 3/6/9 and left 1/4/7) and impose no further restrictions.
    \item \textbf{Immer-Any.} An ideal unconstrained setup where images are selected and paired without any restriction. It represents fully open scenes in the real world.
\end{itemize}

\Cref{tab:dataset-prop} summarizes all protocols by the factors studied.

\subsection{Evaluation Setup}
\label{subsec:protocol-organization}

\begin{figure}
    \centering
    \includegraphics[width=\linewidth]{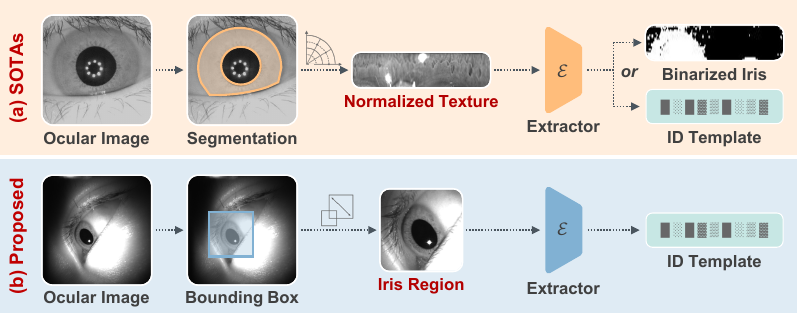}
    \caption{Paradigm comparison between SOTAs and the proposed method. (a) SOTAs segment the iris region and unwrap the iris contour into a rectangular normalized texture. Unreliable normalization yields ill-unwrapped textures, degrading performance. (b) The proposed method dispenses with normalization and employs the cropped iris region, which is more robust and performs better in immersive iris recognition, as later experimentally found.}
    \label{fig:paradigm}
    \vspace{-3mm}
\end{figure}

Following standard practice~\cite{daugman2009iris, ISO19794-6-2011, sheela2010iris}, we pair images for each protocol to support both iris verification and identification tasks, with single-eye and dual-eye testing enabled. 

\noindent \textbf{Verification and Identification.} \textit{Iris verification} refers to 1-to-1 matching, where the system determines if a probe iris matches a stored one. We sample up to 1.5M genuine image pairs and up to 3M impostor pairs following each protocol's design, which ensures reliable estimation of false acceptance rates (FAR) down to 1e-5. \textit{Iris identification} refers to 1-to-N matching, where the system compares a probe iris against all entries in a gallery to determine the probe's identity. We construct a shared gallery with one high-quality sample per class for all protocols except Immer-Dilation (omitted for identification since both of its paired images are assumed degraded), and pair each entry with up to 100 probe samples per class according to the protocol design.

\noindent \textbf{Single-eye and Dual-eye testing.} They refer to two common operation modes: a more convenient one using a single eye, and a more secure one using both eyes jointly for identity determination. For \textit{single-eye testing}, pair formation is carried out only within the same eye side. For \textit{dual-eye testing}, simultaneously captured left- and right-eye images are paired, and acceptance requires both to match. We omit dual-eye testing for Immer-Select, since extreme gaze points from both sides must be removed, resulting in an oversimplified test with only gaze points 2/5/8 remaining.

\section{The Proposed Method}
\label{sec:baseline}

\begin{table}[tbp]

\caption{Verification FRR@FAR ($\downarrow$) of SOTAs trained on CASIA-T and tested on (a) CASIA-T and (b) Immer-Any. Results on left and right eyes are averaged due to space constraints.}
\label{tab:exp-onaxis}

\footnotesize
\centering
\begin{tabular}{clccc}
\toprule
                                    & \textbf{Method}                                               & \multicolumn{3}{c}{\textbf{Performance}}                                                         \\
\multirow{-2}{*}{\makecell[c]{\textbf{Test}\\\textbf{Protocol}}} & FRR@FAR (\%) $\downarrow$                                     & 1e-1                             & 1e-3                             & 1e-5                             \\
\midrule
\multirow{7}{*}{\makecell[l]{\textbf{(a)}\\\textbf{CASIA-T}}} & Gabor~\cite{daugman2009iris}            & 0.36                         & 1.03                         & 5.24                         \\
                                    & OM~\cite{sun2008ordinal}                & 0.24                         & 1.76                         & 5.14                         \\
                                    & Maxout~\cite{zhang2018deep}             & 1.82                         & 17.93                        & 47.26                        \\
                                    & Maxout-BA~\cite{wei2022towards}         & 2.14                         & 21.49                        & 50.38                        \\
                                    & UE-UGCL~\cite{wei2022towards}           & 1.29                         & 11.51                        & 35.20                        \\
                                    & CM~\cite{wei2022contextual}             & 1.50                         & 14.71                        & 38.76                        \\
                                    & ComplexIrisNet~\cite{nguyen2022complex} & 1.08                         & 13.74                        & 35.79                        \\
\midrule
\multirow{7}{*}{\makecell[l]{\textbf{(b)}\\\textbf{Immer-Any}}} & Gabor~\cite{daugman2009iris}            & 32.12                        & 64.33                        & 85.47                        \\
                                    & OM~\cite{sun2008ordinal}                & 30.85                        & 72.18                        & 88.48                        \\
                                    & Maxout~\cite{zhang2018deep}             & 38.83 & 83.61 & 94.09 \\
                                    & Maxout-BA~\cite{wei2022towards}         & 36.43 & 78.73 & 91.94 \\
                                    & UE-UGCL~\cite{wei2022towards}           & 34.62 & 79.02 & 92.42 \\
                                    & CM~\cite{wei2022contextual}             & 38.68 & 78.63 & 90.90 \\
                                    & ComplexIrisNet~\cite{nguyen2022complex} & 42.25                        & 81.07                        & 93.14                        \\
\bottomrule
\end{tabular}

\vspace{-3mm}
\end{table} 
\begin{table*}[tbp]

\caption{Verification FRR@FAR ($\downarrow$) of SOTAs and the proposed method on 4 evaluation protocols of increasing difficulty, capturing multiple challenges in combination. \textbf{Bold} and \underline{underline} indicate the best and second-best results, respectively; the same applies hereafter. Note that the dual-eye testing on Immer-Select is oversimplified hence omitted, as discussed in~\cref{subsec:protocol-organization}.}
\label{tab:exp-1v1-general}

\centering
\footnotesize
\begin{tabular}{ll|ccc|ccc|ccc|ccc}
\toprule
\multirow{2}{*}{\textbf{Eye}} & \textbf{Method}           & \multicolumn{3}{c|}{\textbf{Immer-Control}} & \multicolumn{3}{c|}{\textbf{Immer-Fix}} & \multicolumn{3}{c|}{\textbf{Immer-Select}} & \multicolumn{3}{c}{\textbf{Immer-Any}} \\
                              & FRR@FAR (\%) $\downarrow$ & 1e-1           & 1e-3           & 1e-5           & 1e-1          & 1e-3          & 1e-5         & 1e-1           & 1e-3           & 1e-5          & 1e-1          & 1e-3          & 1e-5         \\
\midrule
\multirow{9}{*}{Left}         & Gabor~\cite{daugman2009iris}                     & 3.26       & 11.06      & 19.14      & 8.33      & 22.44     & 44.12    & 22.22      & 53.43      & 87.66     & 30.75     & 62.60     & 83.20    \\
                              & OM~\cite{sun2008ordinal}                        & 3.09       & 18.10      & 39.16      & 8.09      & 31.28     & 53.37    & 19.67      & 63.36      & 81.77     & 29.51     & 73.13     & 87.81    \\
                              & Maxout~\cite{zhang2018deep}                    & 0.74       & 8.82       & 20.95      & 3.18      & 18.61     & 34.13    & 7.48       & 45.03      & 74.33     & 9.54      & 51.48     & 77.32    \\
                              & Maxout-BA~\cite{wei2022towards}                 & 0.43       & 6.90       & 19.58      & 2.42      & 16.25     & 34.84    & 5.33       & 40.49      & 72.06     & 7.06      & 46.65     & 75.67    \\
                              & UE-UGCL~\cite{wei2022towards}                   & 0.23       & 3.17       & 10.63      & 1.38      & 10.65     & 24.36    & 3.04       & 25.61      & 56.32     & 3.84      & 31.77     & 60.57    \\
                              & CM~\cite{wei2022contextual}                        & \textbf{0.17}       & 1.93       & 7.18       & 1.18      & \underline{7.73}      & 18.35    & 2.52       & \textbf{17.52}      & \textbf{45.03}     & \underline{3.35}      & \textbf{23.39}     & \textbf{49.93}    \\
                              & ComplexIrisNet~\cite{nguyen2022complex}            & \underline{0.19}       & 2.07       & 7.32       & \underline{0.98}      & 7.97      & \underline{19.73}    & \underline{2.23}       & \underline{19.65}      & 49.13     & 3.63      & 27.81     & 57.62    \\
                              & NormKeep                 & 0.25       & \underline{1.70}       & \underline{6.41}       & 1.92      & 9.49      & 19.77    & 4.33       & 23.48      & 49.20     & 5.48      & 28.95     & 56.63    \\
                              & \gcell\textbf{NormFree} (ours)          & \gcell0.21       & \gcell\textbf{1.59}       & \gcell\textbf{5.50}       & \gcell\textbf{0.83}      & \gcell\textbf{6.28}      & \gcell\textbf{15.22}    & \gcell\textbf{2.17}       & \gcell20.99      & \gcell\underline{47.96}     & \gcell\textbf{2.36}      & \gcell\underline{24.03}     & \gcell\underline{52.04}    \\
\midrule
\multirow{9}{*}{Right}        & Gabor~\cite{daugman2009iris}                     & 3.75       & 12.86      & 25.84      & 9.62      & 26.46     & 51.29    & 24.09      & 55.72      & 84.09     & 33.49     & 66.05     & 87.74    \\
                              & OM~\cite{sun2008ordinal}                        & 4.74       & 19.77      & 37.26      & 10.21     & 32.30     & 52.64    & 22.38      & 61.40      & 81.17     & 32.19     & 71.23     & 89.14    \\
                              & Maxout~\cite{zhang2018deep}                    & 1.25       & 11.99      & 37.21      & 3.65      & 21.52     & 39.75    & 8.30       & 47.35      & 79.66     & 12.49     & 54.94     & 80.53    \\
                              & Maxout-BA~\cite{wei2022towards}                 & 0.65       & 9.71       & 33.70      & 2.54      & 18.67     & 37.35    & 6.14       & 42.81      & 76.36     & 9.53      & 50.07     & 76.92    \\
                              & UE-UGCL~\cite{wei2022towards}                   & 0.15       & 3.83       & 14.43      & 1.17      & 10.98     & 26.91    & 2.75       & 26.89      & 59.84     & 4.38      & 33.55     & 62.80    \\
                              & CM~\cite{wei2022contextual}                        & 0.12       & 2.65       & 10.39      & 1.14      & \underline{9.07}      & \underline{20.02}    & 2.06       & \underline{18.79}      & \underline{42.69}     & \underline{3.60}      & \underline{26.37}     & \underline{50.11}    \\
                              & ComplexIrisNet~\cite{nguyen2022complex}            & \textbf{0.10}       & 2.47       & 8.55       & \underline{1.01}      & 9.67      & 23.86    & \underline{1.93}       & 20.47      & 48.14     & 3.97      & 32.24     & 60.72    \\
                              & NormKeep                 & 0.42       & \underline{2.15}       & 8.26       & 2.34      & 10.60     & 20.75    & 4.72       & 22.78      & 46.19     & 6.20      & 30.48     & 53.82    \\
                              & \gcell\textbf{NormFree} (ours)          & \gcell\underline{0.11}       & \gcell\textbf{1.69}       & \gcell\textbf{4.93}       & \gcell\textbf{0.74}      & \gcell\textbf{6.41}      & \gcell\textbf{15.02}    & \gcell\textbf{1.75}       & \gcell\textbf{17.21}      & \gcell\textbf{40.64}     & \gcell\textbf{2.28}      & \gcell\textbf{23.86}     & \gcell\textbf{49.56}    \\
\midrule
\multirow{9}{*}{Dual}         & Gabor~\cite{daugman2009iris}                     & 4.66       & 14.05      & 23.17      & 7.12      & 19.62     & 31.09    & -            & -            & -           & 28.50     & 58.60     & 73.80    \\
                              & OM~\cite{sun2008ordinal}                        & 4.67       & 20.06      & 33.89      & 6.83      & 25.31     & 41.04    & -            & -            & -           & 25.50     & 63.81     & 80.28    \\
                              & Maxout~\cite{zhang2018deep}                    & 0.85       & 9.22       & 22.17      & 1.69      & 13.22     & 28.44    & -            & -            & -           & 6.38      & 40.23     & 73.14    \\
                              & Maxout-BA~\cite{wei2022towards}                 & 0.53       & 6.69       & 19.08      & 1.17      & 10.49     & 24.01    & -            & -            & -           & 4.27      & 34.30     & 66.39    \\
                              & UE-UGCL~\cite{wei2022towards}                   & 0.26       & 2.81       & 10.14      & 0.61      & 5.41      & 16.18    & -            & -            & -           & 2.02      & 19.65     & 48.01    \\
                              & CM~\cite{wei2022contextual}                        & 0.24       & 2.53       & 7.81       & 0.59      & 4.83      & \underline{12.23}    & -            & -            & -           & 1.84      & \underline{16.60}     & \textbf{38.44}    \\
                              & ComplexIrisNet~\cite{nguyen2022complex}            & \underline{0.21}       & \underline{2.18}       & 7.57       & \underline{0.50}      & \underline{4.37}      & 12.81    & -            & -            & -           & \underline{1.65}      & 18.42     & 41.71    \\
                              & NormKeep                 & 0.48       & 2.24       & \underline{5.65}       & 1.44      & 6.85      & 14.14    & -            & -            & -           & 3.85      & 20.95     & 44.69    \\
                              & \gcell\textbf{NormFree} (ours)          & \gcell\textbf{0.18}       & \gcell\textbf{1.16}       & \gcell\textbf{5.17}       & \gcell\textbf{0.48}      & \gcell\textbf{3.17}      & \gcell\textbf{10.61}    & \gcell-            & \gcell-            & \gcell-           & \gcell\textbf{1.24}      & \gcell\textbf{13.29}     & \gcell\underline{40.45}   \\
\bottomrule
\end{tabular}
\vspace{-3mm}
\end{table*}

SOTA iris recognition methods mostly follow Daugman's two-stage paradigm~\cite{daugman2009iris}, where normalization pre-processes ocular images into normalized textures, and feature extraction maps textures into identity templates, as illustrated in~\cref{fig:paradigm}(a). This paradigm achieved remarkable success in earlier years, when ocular images were acquired under control and feature extractor designs were relatively primitive. In such cases, normalization produced relatively invariant textures in advance that facilitated recognition. 

However, normalization becomes quite unreliable in the immersive scenario, as distortion and degradation have been widely reported to yield ill-processed textures~\cite{karakaya2019effect,tomeo2015biomechanical,malgheet2021iris,proencca2009ubiris}, while intra-subject variations further deteriorate texture consistency~\cite{karakaya2021fragile,karakaya2016study}. Though some SOTAs~\cite{wei2022towards,wei2022contextual,nguyen2022complex} attempt to improve normalization design or incorporate post-processing, they remain largely non-intuitive and suboptimal; further see the supplementary material. Meanwhile, DNN-based feature extractors have shown growing capability in handling complex feature representations, largely taking over the role that normalization once played. In this sense, normalization may be becoming a technical debt.

To dispense with the fallible normalization, we reframe an end-to-end recognition paradigm that directly utilizes ocular images for feature extraction. Simply, we detect the iris region via a pre-trained detector, samely as in data cleaning (\cref{subsec:data-cleaning}), and crop it from ocular images with a square bounding box. This process is far more robust than normalization and can be performed efficiently on the fly. The bounding box is expanded by a factor of 1.2 to include contextual cues from adjacent ocular regions and then resized to the desired input shape.

While most SOTAs design dedicated feature extraction modules, we instead adopt a general face recognition architecture~\cite{mi2022duetface,mi2024privacy}. For decades, FR has demonstrated remarkable success in open scenes, owing not to elaborate preprocessing, but rather to robust feature extractors and discriminative training objectives. Accordingly, we employ a ResNet~\cite{he2016deep} of comparable size to SOTAs and train it using an angular-margin-based loss, \textit{wlog.}, ArcFace~\cite{deng2019arcface}. 

\Cref{fig:paradigm}(b) illustrates our proposed paradigm. We refer to this simple yet effective design as \textit{NormFree}, which serves as a baseline method later in the benchmarks to highlight the inherent benefits of being normalization-free.

\section{Benchmarks}
\label{sec:benchmark}

\begin{table*}[tbp]

\caption{Verification FRR@FAR ($\downarrow$) of SOTAs and the proposed method on 4 evaluation protocols of isolated challenges.}
\label{tab:exp-1v1-factor}

\centering
\footnotesize
\begin{tabular}{ll|ccc|ccc|ccc|ccc}
\toprule
\multirow{2}{*}{\textbf{Eye}} & \textbf{Method}           & \multicolumn{3}{c|}{\textbf{Immer-Occlusion}} & \multicolumn{3}{c|}{\textbf{Immer-Dilation}} & \multicolumn{3}{c|}{\textbf{Immer-Light}} & \multicolumn{3}{c}{\textbf{Immer-Gaze}} \\
                              & FRR@FAR (\%) $\downarrow$ & 1e-1            & 1e-3            & 1e-5           & 1e-1           & 1e-3            & 1e-5           & 1e-1             & 1e-3             & 1e-5            & 1e-1           & 1e-3          & 1e-5          \\
\midrule
\multirow{9}{*}{Left}         & Gabor~\cite{daugman2009iris}                     & 12.63       & 30.92       & 43.16      & 3.33       & 11.98       & 21.19      & 6.88         & 20.90        & 37.94       & 22.50      & 53.62     & 73.12     \\
                              & OM~\cite{sun2008ordinal}                        & 8.63        & 28.45       & 45.46      & 4.62       & 16.80       & 24.23      & 7.63         & 36.12        & 65.99       & 21.23      & 68.97     & 88.93     \\
                              & Maxout~\cite{zhang2018deep}                    & 2.45        & 22.90       & 37.54      & 0.93       & 9.34        & 24.23      & 1.69         & 18.09        & 41.86       & 5.10       & 45.20     & 75.98     \\
                              & Maxout-BA~\cite{wei2022towards}                 & 1.68        & 20.70       & 35.07      & 1.02       & 9.00        & 21.56      & 1.28         & 15.30        & 39.74       & 2.98       & 39.14     & 73.71     \\
                              & UE-UGCL~\cite{wei2022towards}                   & 0.63        & 13.33       & 27.66      & 0.32       & 4.14        & 11.71      & 0.50         & 6.40         & 18.67       & 1.28       & 22.13     & 55.52     \\
                              & CM~\cite{wei2022contextual}                        & 0.50        & 7.31        & 16.59      & 0.08       & 6.59        & 12.10      & 0.34         & 5.29         & \underline{16.62}       & \textbf{0.93}       & \textbf{13.25}     & \textbf{38.92}     \\
                              & ComplexIrisNet~\cite{nguyen2022complex}            & 0.58        & 11.67       & 29.58      & \underline{0.04}       & \underline{3.34}        & \underline{8.22}       & \underline{0.20}         & \underline{4.78}         & 21.62       & 1.07       & \underline{15.31}     & \underline{40.31}     \\
                              & NormKeep                 & \underline{0.48}        & \underline{5.01}        & \underline{12.35}      & 0.13       & 5.01        & 13.66      & 0.61         & 7.30         & 20.82       & 1.31       & 16.30     & 46.63     \\
                              & \gcell\textbf{NormFree} (ours)          & \gcell\textbf{0.38}        & \gcell\textbf{2.51}        & \gcell\textbf{8.23}       & \gcell\textbf{0.02}       & \gcell\textbf{2.26}        & \gcell\textbf{4.53}       & \gcell\textbf{0.42}         & \gcell\textbf{2.02}         & \gcell\textbf{9.79}        & \gcell\underline{0.97}       & \gcell16.40     & \gcell45.04     \\
\midrule
\multirow{9}{*}{Right}        & Gabor~\cite{daugman2009iris}                     & 10.74       & 28.40       & 42.55      & 8.87       & 25.46       & 37.90      & 12.27        & 33.00        & 55.17       & 23.61      & 53.64     & 71.72     \\
                              & OM~\cite{sun2008ordinal}                        & 8.50        & 27.47       & 46.61      & 8.62       & 29.60       & 43.44      & 9.44         & 42.12        & 63.49       & 24.95      & 66.18     & 88.05     \\
                              & Maxout~\cite{zhang2018deep}                    & 3.06        & 22.03       & 33.84      & 4.64       & 20.26       & 38.69      & 3.58         & 23.71        & 40.88       & 7.55       & 48.21     & 79.96     \\
                              & Maxout-BA~\cite{wei2022towards}                 & 1.76        & 19.88       & 30.80      & 3.31       & 17.27       & 25.31      & 2.99         & 22.88        & 40.88       & 5.31       & 42.32     & 74.85     \\
                              & UE-UGCL~\cite{wei2022towards}                   & 0.43        & 11.32       & 26.38      & 1.00       & 12.52       & 22.45      & 1.61         & 15.49        & 33.70       & 1.75       & 22.76     & 56.77     \\
                              & CM~\cite{wei2022contextual}                        & \underline{0.42}        & 7.44        & 19.11      & 0.88       & \underline{8.74}        & \underline{17.09}      & 1.13         & 12.36        & 27.82       & 1.01       & \textbf{15.32}     & 40.52     \\
                              & ComplexIrisNet~\cite{nguyen2022complex}            & 0.46        & 12.26       & 27.71      & \textbf{0.21}       & 9.59        & 29.43      & \underline{0.95}         & \underline{10.50}        & \underline{26.39}       & \underline{0.97}       & 16.93     & \textbf{40.25}     \\
                              & NormKeep                 & 0.66        & \underline{5.31}        & \underline{12.43}      & 2.19       & 14.31       & 30.47      & 4.44         & 22.40        & 38.27       & 1.79       & \underline{16.46}     & 42.81     \\
                              & \gcell\textbf{NormFree} (ours)          & \gcell\textbf{0.26}        & \gcell\textbf{2.24}        & \gcell\textbf{5.52}       & \gcell\underline{0.33}       & \gcell\textbf{8.73}        & \gcell\textbf{10.85}      & \gcell\textbf{0.49}         & \gcell\textbf{9.86}         & \gcell\textbf{21.45}       & \gcell\textbf{0.76}       & \gcell17.74     & \gcell\underline{40.41}     \\
\midrule
\multirow{9}{*}{Dual}         & Gabor~\cite{daugman2009iris}                     & 9.07        & 26.02       & 39.54      & 6.77       & 20.77       & 30.55      & 8.87         & 25.90        & 41.84       & 17.77      & 44.61     & 63.70     \\
                              & OM~\cite{sun2008ordinal}                        & 6.48        & 25.66       & 39.21      & 5.45       & 28.25       & 42.09      & 6.65         & 39.03        & 62.44       & 16.94      & 52.04     & 73.66     \\
                              & Maxout~\cite{zhang2018deep}                    & 1.29        & 15.75       & 35.30      & 1.41       & 13.17       & 24.45      & 1.25         & 12.43        & 27.52       & 3.08       & 31.32     & 60.07     \\
                              & Maxout-BA~\cite{wei2022towards}                 & 0.78        & 12.95       & 26.76      & 0.54       & 12.60       & 19.77      & 0.53         & 11.06        & 27.83       & 1.42       & 23.55     & 54.12     \\
                              & UE-UGCL~\cite{wei2022towards}                   & \textbf{0.27}        & 5.10        & 17.22      & 0.27       & 3.27        & 6.53       & \textbf{0.39}         & 5.89         & 16.55       & 0.45       & 9.32      & 34.44     \\
                              & CM~\cite{wei2022contextual}                        & \underline{0.29}        & \underline{3.77}        & \underline{10.53}      & \textbf{0.12}       & \underline{2.62}        & \underline{5.87}       & 0.53         & \underline{4.33}         & 14.99       & \underline{0.33}       & 7.25      & 25.02     \\
                              & ComplexIrisNet~\cite{nguyen2022complex}            & 0.33        & 6.12        & 18.59      & \underline{0.16}       & 3.73        & 13.71      & 0.53         & 4.68         & \underline{12.87}       & 0.43       & 7.06      & \textbf{20.79}     \\
                              & NormKeep                 & 0.67        & 4.99        & 11.73      & 0.54       & 5.14        & 8.13       & 1.57         & 9.43         & 21.35       & 0.91       & \underline{6.49}      & \underline{21.13}     \\
                              & \gcell\textbf{NormFree} (ours)          & \gcell0.31        & \gcell\textbf{1.63}        & \gcell\textbf{3.92}       & \gcell0.28       & \gcell\textbf{1.02}        & \gcell\textbf{1.87}       & \gcell\underline{0.44}         & \gcell\textbf{1.99}         & \gcell\textbf{6.12}        & \gcell\textbf{0.32}       & \gcell\textbf{6.41}      & \gcell23.39    \\
\bottomrule
\end{tabular}
\vspace{-3mm}
\end{table*}

We present a comprehensive benchmark among SOTAs and NormFree on the ImmerIris dataset. \textbf{In short, we find:} 1) ImmerIris effectively captures the unique challenges of immersive recognition; 2) SOTA methods generalize poorly to immersive protocols, indicating the need for methodological advances; 3) Being normalization-free delivers robust performance and suggests a promising future direction.

\subsection{Experimental Setup}
\label{subsec:exp-setup}

\noindent \textbf{Compared SOTAs.} We compare 7 normalization-based SOTAs with the proposed NormFree, including Gabor~\cite{daugman2009iris} and OM~\cite{sun2008ordinal} using hand-crafted filters, and Maxout~\cite{zhang2018deep}, Maxout-BA~\cite{wei2022towards}, UE-UGCL~\cite{wei2022towards}, CM~\cite{wei2022contextual}, and ComplexIrisNet~\cite{nguyen2022complex} using DNN-based extractors. To isolate the effect of waiving normalization, we also ablate NormFree by using the same model architecture but keeping the normalization instead, termed \textit{NormKeep}.

\noindent \textbf{Datasets.} We use CASIA-IrisV4-Thousand~\cite{casia-iris-v4} (CASIA-T), a dataset for traditional iris recognition acquired under control, as a baseline. It contains 20K images from 1,000 subjects. For the immersive scenario, we use ImmerIris.

\noindent \textbf{Metrics.} We report the FRR@FAR of 1e-1/1e-3/1e-5 for verification, and the rank-1 accuracy for identification.

\noindent \textbf{Implementation.} Deferred to the supplementary material.

\subsection{Controlled and Cross-Domain Evaluation}
\label{subsec:exp-controlled}

We first train SOTAs on CASIA-T to establish a performance baseline for traditional iris recognition. \Cref{tab:exp-onaxis}(a) reports the verification FRR@FAR of the trained models on the test set of CASIA-T (training-free methods are evaluated directly). Gabor and OM perform most strongly, followed by DNN-based methods, which are supposed to outperform Gabor and OM on larger datasets~\cite{zhang2018deep,wei2022towards,wei2022contextual,nguyen2022complex} but are partly limited here by CASIA-T's smaller training volume (14,000 images). Overall, all SOTA methods are effective under this controlled setup.


Next, we evaluate the CASIA-trained models on the Immer-Any protocol to examine the cross-domain discrepancy between traditional and immersive data. Note that the normalization is applicable by design to both on-axis and off-axis images, so the inputs are nominally aligned. Thus, the observed performance differences can be primarily attributed to the domain gap between datasets. As shown in~\cref{tab:exp-onaxis}(b), all SOTAs suffer a drastic performance degradation, becoming barely usable. This result attests that immersive iris data pose unique challenges compared with traditional ones, which are captured by ImmerIris.

\subsection{General Verification Performance}
\label{subsec:exp-1v1-general}

\begin{table*}[tbp]

\caption{Identification rank-1 accuracy ($\uparrow$) of SOTAs and the proposed method on different evaluation protocols.}
\label{tab:exp-1vn}

\centering
\footnotesize
\begin{tabular}{ll|YYYY|YYY}
\toprule
\textbf{Eye}           & \textbf{Method}  & \textbf{Control} & \textbf{Fix} & \textbf{Select} & \textbf{Any} & \textbf{Occlusion} & \textbf{Light} & \textbf{Gaze} \\
\midrule
\multirow{9}{*}{Left}  & Gabor~\cite{daugman2009iris}           & 91.99          & 86.44      & 48.52         & 49.40      & 65.72            & 65.57        & 50.01        \\
                       & OM~\cite{sun2008ordinal}              & 85.55          & 79.04      & 45.15         & 44.79      & 45.58            & 62.30        & 47.16        \\
                       & Maxout~\cite{zhang2018deep}          & 97.70          & 93.97      & 76.97         & 76.12      & 77.39            & 81.97        & 79.96        \\
                       & Maxout-BA~\cite{wei2022towards}       & 98.74          & 95.79      & 82.18         & 81.78      & 89.40            & 88.52        & 85.75        \\
                       & UE-UGCL~\cite{wei2022towards}         & 99.35          & 97.27      & 89.92         & 88.83      & 93.64            & 96.72        & 93.25        \\
                       & CM~\cite{wei2022contextual}              & 99.59          & 98.19      & \underline{93.59}         & \underline{92.65}      & \underline{95.41}            & 95.08        & \textbf{96.14}        \\
                       & ComplexIrisNet~\cite{nguyen2022complex}  & \underline{99.67}          & 97.89      & 92.52         & 91.14      & 93.64            & 96.72        & 95.18        \\
                       & NormKeep        & \textbf{99.76}          & \underline{98.26}      & 92.28         & 91.99      & 95.41            & \underline{98.36}        & 94.75        \\
                       & \gcell\textbf{NormFree} (ours) & \gcell99.52          & \gcell\textbf{98.91}      & \gcell\textbf{93.87}         & \gcell\textbf{94.39}      & \gcell\textbf{98.23}            & \gcell\textbf{98.36}        & \gcell\underline{95.49}        \\
\bottomrule
\end{tabular}
\vspace{-3mm}
\end{table*}

In~\cref{subsec:exp-1v1-factor,subsec:exp-1v1-general,subsec:exp-1vn}, we train SOTAs and NormFree on ImmerIris and conduct comprehensive benchmarks across the 8 challenge-capturing protocols (\cref{subsec:protocol-rationale}). We first evaluate the general verification performance on 4 protocols that combine multiple challenges as representing different operation modes, in increasing difficulty: Immer-Control, Fix, Select and Any. From~\cref{tab:exp-1v1-general}, we highlight several findings:

ImmerIris is the largest iris dataset to date that provides a training set 25$\times$ larger than CASIA-T. DNN-based SOTAs trained on ImmerIris perform much better on Immer-Control than on CASIA-T in~\cref{tab:exp-onaxis}(a) (their setup differ only by off-axis distortion), showing that performance can benefit from increased data volume. Nonetheless, all SOTAs degrade sharply as the protocols incorporate more variations and distortions, reflecting their challenging nature.

Gabor and OM perform poorly on all protocols. Since they are training-free methods whose effectiveness primarily depends on normalization, this suggests the isolated ineffectiveness of normalization even in the simplest immersive operation mode. On the other hand, even the best SOTA performs unsatisfactorily on the two most challenging protocols (Select/Any), suggesting that methodological advances are necessary to meet immersive demands.

The proposed NormFree performs surprisingly well, ranking first or second in almost all cases despite its simple design. This suggests the potential of dispensing with normalization. On the other hand, the ablated NormKeep usually falls behind the best SOTAs, and exhibits a considerable gap from NormFree, demonstrating the stand-alone benefit of being normalization-free.

\subsection{Verification on Isolated Challenges}
\label{subsec:exp-1v1-factor}

In~\cref{tab:exp-1v1-factor}, we then analyze how isolated factors (\ie, occlusion, dilation, illumination change and gaze variation) affect recognition performance on their respective protocols:

Though occlusion partly obscures iris textures, it affects performance only moderately, as FRR@FAR(1e-5) on Occlusion increases by 9.34\% on average compared with Control. Surprisingly, dilation does not substantially degrade recognizability. However, SOTAs' performance drops considerably on Light, where paired irises naturally dilate and constrict due to illumination variation, suggesting that normalization is also sensitive to lighting changes. Gaze variation accounts for an average 36.99\% degradation, making it the most challenging factor in the immersive scenario.

The proposed NormFree handles occlusion and dilation effectively and is even robust to illumination changes, ranking first or second across these protocols. Its performance gain over NormKeep is more significant here (4.12–16.82\%) than in~\cref{tab:exp-1v1-general}, suggesting that waiving normalization may be especially beneficial for handling degradations. Nonetheless, though NormFree outperforms most SOTAs under gaze variation, it lacks a decisive advantage, suggesting that dedicated improvements may be required to address this factor, which we will take up in future research.

\subsection{Identification Performance}
\label{subsec:exp-1vn}



We report identification results in~\cref{tab:exp-1vn}. Due to space constraints, we show only left-eye results here and defer the rest to the supplementary material. Overall, we observe similar trends as in verification: immersive iris recognition introduces distinct challenges from distortion, variation, and degradation, leaving substantial room for improvement for current SOTAs. Dispensing with normalization yields a clear performance gain, which merits further study.





\subsection{Ablation Studies}
\label{subsec:exp-abal}
\begin{table}[tbp]

\caption{Verification FRR@FAR ($\downarrow$) of NormKeep and NormFree under alternative (Alt.) model scale and iris normalization technique, averaged over left and right eyes on Immer-Any.}
\label{tab:exp-abal}

\footnotesize
\centering
\begin{tabular}{llccc}
\toprule
\multirow{2}{*}{\textbf{Method}} & \textbf{Settings}          & \multicolumn{3}{c}{\textbf{Performance}} \\
                                 & FRR@FAR (\%) $\downarrow$ & 1e-1          & 1e-3           & 1e-5          \\
\midrule
\multirow{3}{*}{NormKeep}         & \textbf{Default}                   & 5.84      & 29.72      & 55.23     \\
                                 & Alt. Model                & 6.08      & 31.51      & 55.41     \\
                                 & Alt. Normalization                 & 3.25      & 24.08      & 52.77     \\
\midrule
\multirow{2}{*}{NormFree}         & \textbf{Default}                   & 2.32      & 23.95      & 50.80     \\
                                 & Alt. Model                     & 2.11      & 24.59      & 51.99     \\
\bottomrule
\end{tabular}
\vspace{-3mm}
\end{table}

We study two factors beyond normalization in~\cref{tab:exp-abal}:

\noindent \textbf{Model scale.} We replace the IR-50 backbone with a smaller IR-18, which is comparable in size to models in recent SOTAs~\cite{nguyen2022complex,wei2022contextual}. NormFree still outperforms NormKeep and surpasses the SOTAs (in~\cref{tab:exp-1v1-general}) with IR-18, suggesting that its performance gain is less variant to model scale.

\noindent \textbf{Normalization technique.} We use a more adaptive implementation of the normalization stage~\cite{wang2022human}, which turns out to improve the performance of NormKeep only marginally. This suggests that the drawback of normalization is general rather than specific to a particular implementation.


\section{Conclusion}
\label{sec:conclusion}

This work introduces ImmerIris, a large-scale public iris dataset dedicated to immersive iris recognition, capturing the unique challenges of perspective distortion, intra-subject variation, and quality degradation. Through extensive benchmarks, we show that SOTAs cannot be readily generalized from controlled setups, underscoring the need for methodological advances. We propose a simple yet effective normalization-free paradigm, which achieves robust performance and points to a promising research direction.

\section*{Acknowledgements}
\small{The computations in this research were performed using the CFFF platform of Fudan University. This research was supported in part by the National Natural Science Foundation of China (Grant No.~62506080) and in part by the China Postdoctoral Science Foundation (Grant No.~GZC20251072). We thank all the volunteers who contributed to the dataset collection. We also thank Jianze Wei (IME, CAS) and Caiyong Wang (BUCEA) for their support in accessing existing datasets and recognition methods. In addition, we thank Yuge Huang, Jianqing Xu, and Jiazhen Ji for helpful discussions.}

{
    \small
    \bibliographystyle{ieeenat_fullname}
    \bibliography{main}
}


\end{document}